# Detection of fraudulent financial papers by picking a collection of characteristics using optimization algorithms and classification techniques based on squirrels

Peyman Mohammadzadeh germi [1], Mohsen Najarbashi [2]
[1]Department of Computer Engineering, Ardabil Branch, Islamic Azad University, Ardabil, Iran
Email: payman.mohammadzadeh@gmail.com
[2]Department of Computer Engineering, Ardabil Branch, Islamic Azad University, Ardabil, Iran
Email: mohsennajarbashi@gmail.com

**ABSTRACT:**
To produce important investment decisions, investors require financial records and economic information. However, most companies manipulate investors and financial institutions by inflating their financial statements. Fraudulent Financial Activities exist in any monetary or financial transaction scenario, whether physical or electronic. A challenging problem that arises in this domain is the issue that affects and troubles individuals and institutions. This problem has attracted more attention in the field in part owing to the prevalence of financial fraud and the paucity of previous research. For this purpose, in this study, the main approach to solve this problem, an anomaly detection-based approach based on a combination of feature selection based on squirrel optimization pattern and classification methods have been used. The aim is to develop this method to provide a model for detecting anomalies in financial statements using a combination of selected features with the nearest neighbor classifications, neural networks, support vector machine, and Bayesian. Anomaly samples are then analyzed and compared to recommended techniques using assessment criteria. Squirrel optimization's meta-exploratory capability, along with the approach's ability to identify abnormalities in financial data, has been shown to be effective in implementing the suggested strategy. They discovered fake financial statements because of their expertise.

## 1. INTRODUCTION

According to the current state of the global economy, fraud prevention and detection activities are on the rise [1]. This is a reaction to the rise in fraudulent activities over the previous many years. In recent times, as the quantity of data needed for the analysis has grown, data mining technologies have become more popular, and tracking suspicious transactions has emerged as the most efficient means of identifying fraud [2, 3] .

The NBA, Dutch Professional Organization of Accountants, says that auditing the financial statements of medium and large enterprises is an essential part of the accountants' job in avoiding fraud. An audit is performed in order to make sure that financial statements are free of serious misrepresentation due to fraud or accidental mistakes. In its protocol, the NBA says that fraud is "willfully making a false statement to gain an unfair advantage [4]." On the other hand, given that fraud in financial statements is very uncommon but that losses sustained as a result of fraud account for a significant portion of the total losses experienced by financial institutions, the identification of fraud as a floor anomaly is quite important. The realm of finance is brought into the discussion. A key part of data mining is anomaly detection, which looks for strange or unexpected information in a set of data. Data mining and machine learning have been extensively explored in the study of anomaly detection in financial reporting because they are commonly used to extract and expose hidden truths from vast volumes of data, which is critical in fraud detection. Significant but unusual occurrences may be indicated by anomalies, which can lead to crucial actions in a broad variety of fields. Disturbing patterns in a credit card transaction might point to fraud [5, 6] .

In light of the fact that in the area of fraud detection, greater emphasis is placed on the categorization of labeled data, and in light of the fact that fraud is not only multifaceted but also continually evolving, classification-based models might be seen as a viable option. The financial data set is uneven since the number of fraudulent transactions is modest relative to the number of legitimate transactions. Consequently, anomaly detection techniques are a viable option for spotting financial fraud data [7, 8].

On the other hand, owing to the rising diversity of fraudulent tactics in health insurance, the traits and characteristics of counterfeit samples compared to those of regular samples have grown significantly. If the input dataset includes a significant number of irrelevant and related characteristics, the computing overhead of the machine learning approach will grow, and the classification performance will be less accurate. The attribute selection procedure chooses a subset of the most important variables or characteristics that may accurately reflect the original data. This subset may help find the data's pattern quickly and accurately so that actionable intelligence can be gained [9, 10].

This has been discussed by a great number of authors in the field of identifying anomalies with data mining methods in the financial field. Therefore, in this study, a mixture of the following subset of characteristics based on the squirrel optimization technique and closest neighbor classification algorithms, neural networks, support vector machines, and Bayesian to identify fraud has been used in financial statements. The experiments are completely based on using the standard financial data set to identify fraud detection patterns in financial statements and for evaluating the model from the same data set. Because of the variety and quantity of characteristics in the data collection, the selection of a subset of features to increase the classification accuracy of the model may be required in the suggested technique. The squirrel optimization technique is employed in this approach to choosing the subset of fundamental characteristics that are critical in detecting fraud in financial accounts. Classification models are evaluated based on their ability to reduce features and errors. The output stage selects the attributes with the least restrictions, and as a consequence, classification algorithms are applied. Financial statements are analyzed and assessed for fraud or normality based on the patterns revealed by categorization techniques, which are used to spot fraud before it happens. The meta-exploratory ability of squirrel optimization should let this method find the most important parts of financial statements, put them into groups, and make accurate predictions based on these cases.

The remainder of this article may be found below. In the next section, we'll take a closer look at the research methodology. The suggested method's specifics will be discussed in length in the third part. The suggested method's implementation and assessment will be discussed in the fourth part. Conclusions will be drawn in the fifth portion of this paper.

## 2. BACKGROUND INFORMATION FOR THE STUDY

### 2.1 REVIEW AND CONFIRM FINANCIAL REPORTS

Organizations and their financial reporting systems are audited independently and objectively by an auditing firm. A company's financial state and performance may be seen in its financial reports. Investors, banks, and suppliers all rely on this data when making economic choices. Accounting companies provide the service of auditing financial statements. Publication of yearly financial accounts is mandated by law for all businesses, big and small. Large and medium-sized businesses need to have their financial records checked by a neutral third party.

The primary goal of financial audits is to guarantee that the financial statements are accurate and detailed enough for investors to have to trust them. An audit by Price Waterhouse Coopers is required to be sure that a report's content is accurate. Before releasing their yearly financial results, companies must get an audit report. Shareholders need an audit report before they can feel secure in their company's financial performance. Whether through fraud or accidental mistake, the financial accounts are devoid of substantial misappropriation, according to this assessment. When a financial statement is described as "clean," it indicates it is devoid of fraud or accidental mistakes, and this is the responsibility of competent and experienced auditors who organize the review of financial statements in accordance with auditing standards. Individually or collectively, financial fraud and abuse may impact the economic choices that consumers make based on their financial records [11]. There are several stages of financial auditing shown in Figure 1.

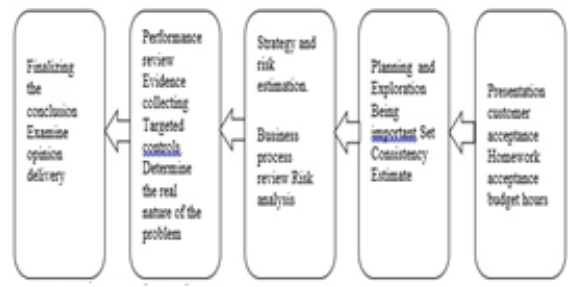

**Fig.1.** MEC network configuration with caching capability

The financial statement is the format for auditing procedures. The first two processes, preparation, planning, and investigation, lead to the selection of the audit company and emphasize consumer acceptability. In the third phase of strategy and risk assessment, auditors identify risks using their knowledge of the company's business, industry, and operating environment. A comprehensive audit strategy is created to address the risks of substantial misrepresentation in the financial statements. This incorporates a trial-and-error approach to numerous financial statement elements.

During the audit, the auditor gathers information by verifying the company's internal audits, monitoring quantities and revealing inventories in financial statements, and acquiring external third-party documents. The inventory of some things, such as the company's financial position, may be independently verified. The following are some of the fundamental tests that may be performed as part of an audit:

- Physical asset inspections, such as inventories or asset counts.
- Examine the records of the transactions.
- Obtain third-party permission, such as from suppliers and customers.
- Study financial statements for information such as pricing comparisons based on indicators from other countries' markets.

Eventually, the auditor comes to a decision and documents his or her findings in the auditors' report [12].

### 2.1.1 ANALYZE THE DATA IN THE FINANCIAL REPORT

Data analysis is an important part of financial statement evaluation. It is possible to examine data in a variety of ways to find out whether there are any abnormalities in the data that were either purposeful or accidental. It is quite difficult to identify irregularities in financial data transactions. Some of the most common ways to study ratios and transactions are through data modeling, sampling, and data analysis.

The primary data is used to find abnormalities in financial statements (such as gross profit and net profit) and ratios (such as long-term debt to capital and reserves) that are reviewed and identified in order to identify forgeries. Anomalies and fraud in financial accounts may be caused by changes to management estimations, accounting formulae, and the inaccurate identification of revenue and spending. The researchers looked at financial records from the technology, health care, and financial services industries. They found that the most common types of fraud are misidentifying income and overstating assets. Researchers have highlighted the following as reasons for fraud: inflationary economy, corruption, misguided executive incentives, unreachable market expectations, pressure from large creditors, legislation, and auditing companies' opportunistic conduct. The following are examples of financial data: sales orders, invoices, purchase orders, shipping papers, and credit card payments. Additionally, transactional data is an important component of stratified data [13, 12].In the context of financial audits, the "set of transaction data" refers to all of the information that pertains to the financial accounts of an organization for a certain fiscal year. These contain accounts in the general ledger, daily ledgers, bank accounts, and data about any applicable taxes, customers, and prospective subsidiaries. The accounts in the general ledger may be separated into two distinct groups: the accounts for the balance sheet and the accounts for the profit and loss. The daily ledger is where a corporation keeps track of the chronological order of all of the business transactions that have taken place inside the organization. This ledger also details the debt and balance associated with each transaction. Benford's law is a way of assessing financial data that is often used for the purpose of conducting financial audits. Benford's law is used in the field of economics to analyze financial data in order to identify anomalies and potential instances of fraud. Benford discovered that the initial digit in several different numerical data sets did not follow a uniform distribution [14].

The solution proposed here addresses only the finding of out-of-reach spots or anomalous financial data sets by using clustering techniques, which is an unsupervised technique. Anomalies can be caused by fraud or by mistakes made by accident, and it is the auditor's job to figure out if the anomaly is a sign of fraud [15].

### 2.1.2 IDENTIFY DEVIATIONS IN FINANCIAL REPORTING

Companies that knowingly deceive investors and creditors by using fictitious financial statements are guilty of committing the offense of misleading financial statements (FSFs). The company's condition, performance, and cash flows are all discussed in the financial statements that are provided. Disclosure of abnormalities is accomplished by the differentiation in the financial statements between normal and questionable patterns of behavior, which permits the creation of more focused solutions. Data discovery, or data mining, is a typical technique for discovering suspicious activity in large quantities of data. Data mining is the discovery of undiscovered patterns in a database and the construction of predictive models utilizing this knowledge. There are two sub-branches of methods for identifying computational anomalies: with and without supervision. One of the primary concerns has always been which procedure is most suitable [16] [17].

## 2.2 ANOMALY DETECTION

A procedure that identifies instances of "abnormal" behavior is known as "anomaly detection." The critical questions in this situation are "how to discover the anomaly?" "how to mitigate its effects." and "how to analyze the problem?" Finding missing points or anomalies in data collection is the primary emphasis of the area of non-monitoring anomalies, which takes an unsupervised approach to the task. One of the first definitions of an anomaly was provided by Grobes, who said that "data that seems to be farther away from the sample data set might be classified as anomalous data." Normal situations are uncommon, according to Goldstein and Ochida's alternative definition of anomaly. "Approaches for identifying anomalous data may be split into supervised and unsupervised methods [18, 19].

### 2.2.1 SUPERVISED METHODS

This indicates that an anomaly detection model is trained using a data set in which the data is tagged and the anomalous data has been pre-identified. (21). Anomalies in data sets may be identified using trained algorithms. Freshly applied in learning, they categorize new data by comparing it to previously classified data. Classification methods include support vector machine (4 SVM), decision resolution (5DT), closest neighbor (6KNN), regression (7LR), neural network (8ANN), and others [20, 21].

**2.2.2 UNSUPERVISED METHODS**

Unsupervised data detection approaches instruct data-driven models in which the anomalous points are unknown, i.e., the features of the aberrant data are unknown and are hence referred to as unlabeled. The unsupervised anomaly identification approach clusters data based only on the inherent properties of the data set. Unsupervised learning is distinguished from supervised learning by the fact that it attempts to uncover the underlying structure in unlabeled data. Similar to zero-day assaults in the realm of intrusion detection, its diagnosis is often seen as an unsupervised learning task owing to the absence of previous information about the nature of distinct abnormalities. Additionally, there is an abundance of unlabeled data but a dearth of labeled data.

Clustering is an unsupervised learning technique that groups comparable data samples without the need for labeled data. Depending on the clustering criteria used, a particular data set is segmented differently by the clustering process. There are several sorts of clustering algorithms; here, we will discuss those used to discover abnormalities. In unsupervised learning, it is thus unnecessary to label the data. In these approaches, data are grouped depending on the distance or density between the data points Hierarchical clustering, k-means clustering, and DBSCAN clustering are examples of clustering algorithms [22, 23].

**2.3 PREVIOUS METHODS**

This section presents a review of prior studies on the models as well as evaluates the numerous anomaly detection techniques in the table that follows. Based on data mining, this study investigates approaches (such as regression, decision trees, neural networks, and Bayesian networks) for identifying financial reporting fraud and proposes an active detection model. Obviously, in this respect, the author should do more study on the creation of active models, the results of which employ a range of clustering approaches in various ways and in which the sole data mining method is the independent clustering technique [24, 25].

Glancy et al. have presented a computer model (CFDM) for the identification of fraudulent activity in order to identify fraudulent financial statements. In this scenario, the act of manipulating information in yearly records filed with the United States Securities and Exchange Commission (SEC) is seen as fraudulent activity. This model was able to identify fraudulent activity in text documents by using maximum expectation as well as hierarchical clustering. The model performed a few repeated clusters on the text documents in order to produce two distinct clusters, one of which includes fictitious information, while the other cluster is composed of genuine information. Using the sign test, clustering results were evaluated and there were only three false positives out of sixty good sets of documents [26].

Turgo et al. used a hierarchical clustering approach, a method was proposed to identify tool-based fraud. Fraud activity is based on the idea that data related to illegal activities are not clustered with other data [27].

In a separate study, the machine learning approaches for fraud detection in automated bank lending systems and the SVM methods, 11BRE-DRSA and 10DRSA, are analyzed. The strategy used to categorize imbalanced data was chosen. The primary cause is the potential classification of uneven data. [28, 29].

In a separate essay, he studies clustering based on collective intelligence algorithms and assesses the method's many quantitative and qualitative characteristics. Using IEEE Xplore, ACM, Science Direct, and Springer standard data sources, the research analyzed 161 papers from 2012 to 2017. These algorithms include the optimization of squirrels, ant colonies, and fish swarms (AFSA). This study demonstrates that for collective intelligence algorithms to be efficient in clustering, it is required to integrate with other clustering approaches such as hierarchical clustering (K-Means and Fuzzy C-Means) [30, 31].

A typical system for detecting financial statement fraud consists of many control layers, each of which may be conducted automatically or under human supervision. Included in the automated layer are machine learning algorithms that produce prediction models based on tagged interactions. In the last decade, significant machine learning research has led to the development of supervised, unsupervised, and semi-regulated algorithms for detecting fraud in financial and credit institutions. In this study, a selection of feature subsets based on squirrel optimization techniques and classification methods were employed to identify abnormalities and fraud in financial accounts. Feature selection in fraud detection systems may improve the accuracy of financial statement fraud detection. It is thus one of the fundamental milestones in this discipline. In the rest of this section, we'll talk about the algorithm for squirrel optimization and then describe the suggested approach.

**3. THE SQUIRREL OPTIMIZATION ALGORITHM**

When flying squirrels start looking for food, the search process starts. During warm weather (fall), squirrels look for food by gliding from tree to tree. In doing so, they alter their position and explore various woodland regions. Since the temperature is warm enough, they can satisfy their daily energy requirements with a diet of several seeds, therefore they swallow them as soon as they discover them. After completing their daily energy requirements, they search for the finest winter food source (each nut). Storing each nut allows them to retain the energy they need to survive in very hard conditions, hence boosting their chances of survival. The lack of leaf cover in deciduous woods during the winter increases the chance of being hunted, therefore animals are less active but do not hibernate. At the conclusion of winter, flying squirrels resume their activity. This repetitive process, which lasts throughout the whole of the flying squirrel's life, is the foundation of the squirrel search algorithm. The purpose of the following assumptions is to simplify the mathematical model:

1. There are two flying squirrels in a deciduous forest, and one squirrel is assumed to be on a tree.
2. Each flying squirrel looks for food on its own and makes the best use of the food resources available by changing the way it looks for food.
3. In the forest, there are only three types of trees, including the common tree, the oak tree (the food source of oak nuts), and the flowering tree (the food source of Kharkivi nuts).

It is anticipated that the studied forest has three oak tree stands and one Khark tree. The number of squirrels is equal to 50, hence n = 50. There are 4 nutrition sources (Nfs) consisting of 1 nut tree and 3 oak kernels, but 46 trees have no Nfs. 92 percent of the squirrel population consists of conventional trees, while the remainder consists of food sources. In winter, however, the number of food sources might change based on the 1 <Nfs <n that fs∈ℤ> 0 limits with a favorite food source[33].

We will now demonstrate features based on the squirrel optimization algorithm and classification techniques on an anomaly and fraud detection system in financial statements based on a combination of subset selection and classification methods. The proposed technique would employ the training data set derived from the standard fraud detection data set to identify fraud detection patterns in financial statements, and the test data set derived from this data set to assess the model. Given the variety and number of characteristics in financial statements, the suggested strategy seems to need the selection of a subset of features in order to improve model classification accuracy. The goal of choosing a subset of characteristics is to get rid of unnecessary features, attributes, and plugins. This makes it possible to improve classification accuracy, classify data quickly and cheaply, and reduce data dimensions, system execution time, and spatial complexity.

In general, having irrelevant and redundant characteristics as a result of addressing the classification issue has an impact on classification performance, particularly with high-dimensional data sets, and diminishes classification accuracy. Furthermore, a high number of features leads to over-training and a poor capacity to generalize to new data and identify test samples. So, because there are so many firms and each one has its characteristics and ways of presenting financial statements, finding fraud in financial statements may become a hard classification problem with less than ideal results.

The goal of picking features is to reduce classification complexity while increasing classification accuracy by selecting relevant information. A goal for optimization exists in the task selection elements of a goal. The main goal of feature selection is to identify the greatest feature combination for optimal classification performance, independent of training cost or feature count. Extra-exploratory feature selection changes the work of feature

## 4. SUGGESTED METHOD

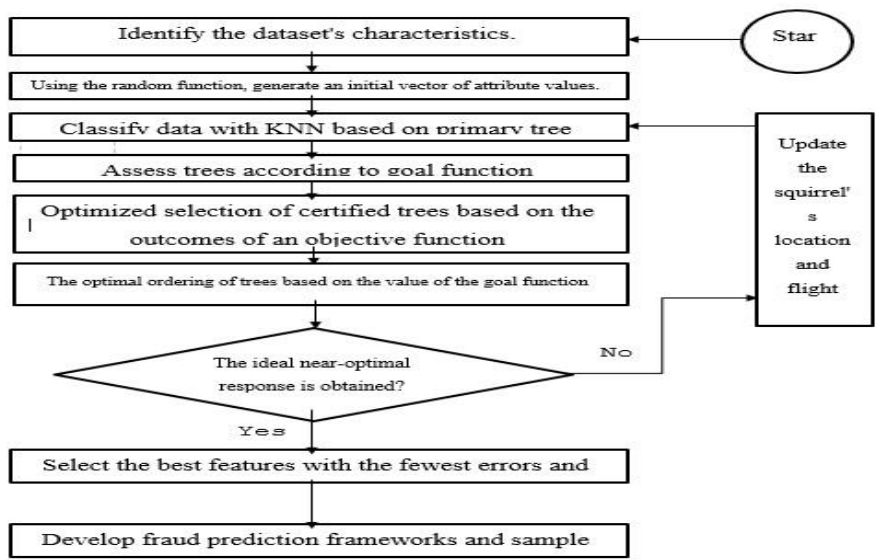

**Fig.2** Flowcharts of the proposed method

selection into a multi-objective optimization problem. The goal is to find a balance so that multiple goals can be optimized. The aims of this optimization approach in order to pick a subset of features will include minimizing the number of features, improving financial statement classification accuracy, and improving test sample prediction accuracy. In reality, the objectives involve two groups: lowering the number of characteristics and improving classification performance. As a result, the solution to the feature optimization issue is a collection of dominating solutions, each of which is a vector consisting of two components: the number of features and the degree of classification error. The purpose of treating the feature selection issue as a minimization problem is to reduce the number of irrelevant features while also lowering the classification error rate. The suggested approach will be developed further in the following sections.

### 4.1 Problem formulation

Similar to other population-based search algorithms, the squirrel search method begins with the random position of flying squirrels. The vector displays the position of the flying squirrel in the d-dimensional search space. Consequently, flying squirrels are able to fly in 1D, 2D, 3D, or more search spaces and modify their position vectors. There are n flying squirrels (FS) in a forest, and each flying squirrel's position can be calculated using a vector. The following matrix displays the locations of all flying squirrels:

$$FS = \begin{bmatrix} FS_{1,1} & FS_{1,2} & \cdots & \cdots & FS_{1,d} \\ FS_{2,1} & FS_{2,2} & \cdots & \cdots & FS_{2,d} \\ \vdots & \vdots & \vdots & \vdots & \vdots \\ \vdots & \vdots & \vdots & \vdots & \vdots \\ FS_{n,1} & FS_{n,2} & \cdots & \cdots & FS_{n,d} \end{bmatrix} \quad (1\text{-}2)$$

Where $Fs_{i,j}$ represents j the flying squirrel's dimensions. Using Equation 2, the starting locations of each flying squirrel in the forest are decided.

$$FS_i = FS_L + U(0,1) \times (FS_U - FS_L) \quad (2\text{-}2)$$

For each dimension of the flying squirrel, $Fs_L$ and $Fs_U$ is the lower and upper bounds of the flying squirrel, which are random numbers distributed in the range [0, 1].

### 4.2 Implementation of the proposed method

In order to find out which aspects of the set of financial statements are the most significant, the technique to feature selection that is provided in this work and is based on an optimization algorithm called the squirrel algorithm is being taken into consideration. The algorithm for squirrel optimization starts with the initial solutions, which are ideas and ways of dealing with the problem that already exists. It then picks the best ones and gets rid of the ones that aren't the best. The aims are merged in this approach, and in the end, we obtain two broad features in the form of a minimization. When evaluating feature subsets, we take into account both the overall goal of decreasing the total number of features as well as the overall goal of lessening the overall amount of classification error. The fitting function is stated as the following equation in order to assess the starting population, pick the expert community, and locate the particles with the greatest weight. This is done in order to discover the particles with the highest weight.

$Minimize F(x) = f_1(x) = L/AL \in A. A \in \mathbb{R}_{++} f_2(x) = 1-(FP+FN/P+N) . (P+N) \in \mathbb{R}_+$

Let L, the number of chosen characteristics from the data collection, and A, represent the entire number of attributes. In the proposed technique, the attributes chosen at each step are utilized to assess the KNN algorithm for classifying training samples in order to calculate the classification error for each particle and acquire the classification error. In each step, the ideal particles are chosen based on the number of specified attributes and ranked by how much better they are. By turning these characteristics from each solution into feature vectors, KNN will attempt to distinguish between legitimate and fraudulent policyholders and establish a margin of safety between the two groups. Finally, the solution with the lowest error rate and the smallest number of features taken from the training data set is determined to be the optimal one, and the extracted features serve as the classification pattern. Figure 4 shows a flowchart of the method suggested, and Figure 5 shows the values of the fit function for the best solution when the squirrel optimization algorithm is used.

As can be seen in Figure 3, the correct solution performance of the solution in the method for squirrel optimization begins at 1 and rapidly lowers in each step until it is repeated until 100 steps are discovered. This process is continued until the optimal solution is found. The best possible result may be attained by using the ratio's function value, which is 0.1103. As can be seen in Table 1, only ten of the sixteen characteristics that are included in the primary data set As a consequence of this, the optimum solutions are progressively created throughout the phases of the squirrel optimization method in order to locate the solution that has the fewest amount of characteristics and the smallest amount of inaccuracy.

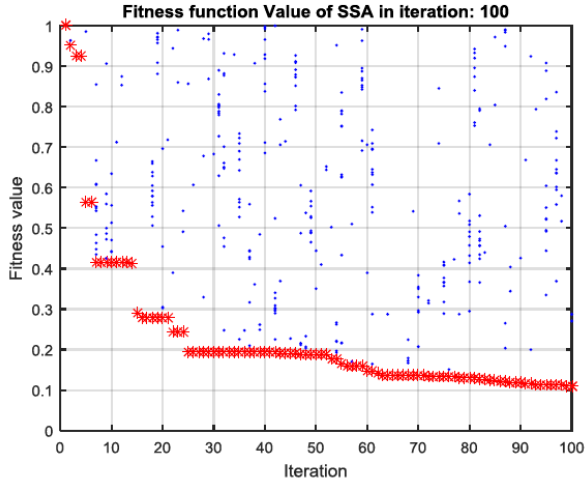

**Fig 3**. Values of the terms of percentage function while the squirrel optimization algorithm is running

The rest characteristic characteristics included in the primary dataset are detailed in Table 1.

**Table 1.** displays the characteristics chosen by the squirrel optimization method.

| Title of the Attributes | $V_1$ | $V_4$ | $V_7$ | $V_9$ | $V_{11}$ | $V_{12}$ | $V_{14}$ | $V_{15}$ | $V_{16}$ |
|---|---|---|---|---|---|---|---|---|---|
| Attributes numbers | 1 | 4 | 7 | 9 | 11 | 12 | 14 | 15 | 16 |

have been chosen as candidates for the role of effective features in the detection of fraudulent financial statements. The resulting feature subset contains the fewest total features and the fewest total errors across all of the sets of training data that comprise the financial statements set. When the closest neighbor method is used, the value of the accuracy of the best solution for the chosen characteristics is found to be 91.8%.

### 4.3 Evaluation of the proposed method

Following the execution of the suggested strategy, which is predicated on the feature selection based on the squirrel optimization algorithm, fresh fraudulent samples can be anticipated in the collection of the financial statements. It is feasible to investigate patterns in which fresh samples are reviewed based on the characteristics picked in the previous stage using feature selection techniques and classification methods. By doing so, it is possible to identify whether or not the financial statements have been falsified. In order to improve the validity of the suggested model for estimating novel fraudulent financial statements obtained from the proposed method and attributes chosen by the squirrel optimization algorithm, the examination was done using the matching class of the latest fraudulent financial statements and the real status of these samples. This was conducted in order to know the efficiency of the suggested method for predicting new fraudulent financial statements based on the proposed method. In order to assess the suggested technique, first, compare the predicted class to the new financial statements and then extract the turbulence matrix parameters based on this comparison. Accuracy, sensitivity, precision, and Criterion F are some of the assessment criteria that may be derived from the turbulence matrix. This is because the turbulence matrix is a common way to measure how well two-layer data can be used to classify things.

Assessment methods obtained from turbulent matrices variables are utilized as a mechanism for the purpose of measuring the quality of the suggested technique and comparing it to other techniques. As a result, in order to assess the suggested technique in this study, we analyze the mixture of feature selection techniques based on squirrel optimization algorithm and closest neighbor algorithms, support vector machine, neural network, and Bayesian algorithm. A comparison of the accuracy of fraud detection using attribute selection based on the squirrel optimization algorithm and classification algorithms is shown in Figure 4. This comparison looks at fresh financial statements.

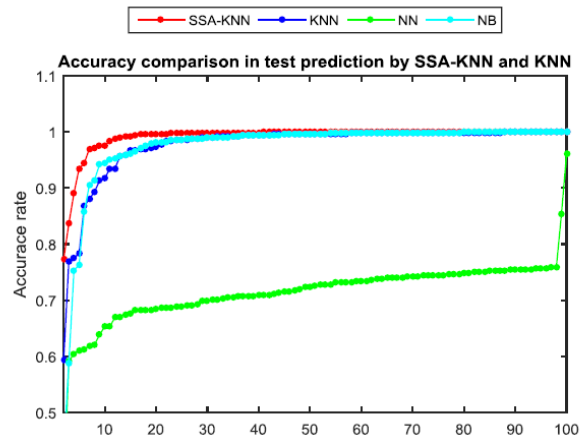

**Fig.4** A comparison of the accuracy of the suggested strategy based on several categorization algorithms.

As shown in Figure 4, the accuracy graph for the suggested method, which is based on a set of feature extraction based on the squirrel optimization algorithm and the nearest neighbor algorithm, is compared to other classifications that do not use the feature selection method to estimate. The accuracy graph for the suggested protocol is derived from a combination of feature selection based on the squirrel optimization algorithm and the nearest neighbor algorithm. The collection is updated with fresh instances of counterfeit items. A comparison of the suggested technique and the classification algorithm is shown in Figure 5. This comparison examines the sensitivity of fraud detection in newly counterfeited samples.

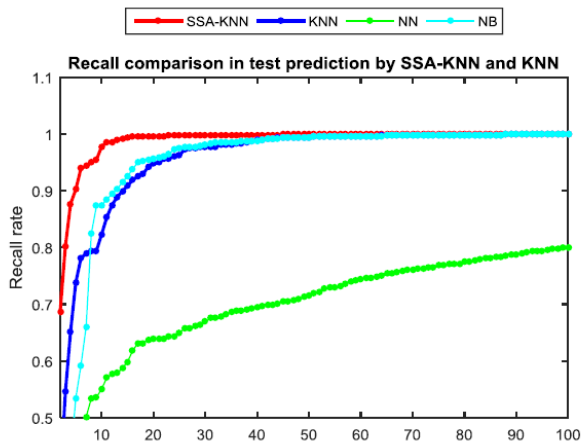

**Fig.5** A Comparison of the Sensitivities of Several Classification Algorithms and the Suggested Technique

The drawn sensitivity criterion chart for the suggested protocol, which is based on a set of attribute selections based on the squirrel optimization algorithm and the nearest neighbor algorithm, especially in comparison to other categories that do not use the attribute selection method to estimate new fake samples in the Social Security Administration database. Figure 8 shows how the suggested method and the classification algorithm compare in terms of how well they can find new fake samples.

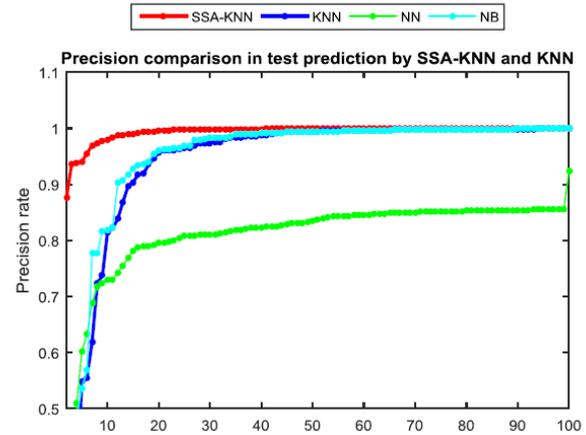

**Fig. 6** Comparison of the precision of the suggested classification algorithm-based technique

If you don't use the attribute selection strategy to forecast new fake samples, you saw in Figure 6 that the precision chart for attribute selection methods based on the squirrel optimization algorithm, the closest neighbor algorithm, and other classification algorithms may be generated. As a last point of interest, Table 2 presents a comparison of the average scores of the assessment criteria for integrating attribute selection based on the squirrel optimization method with classification algorithms for fresh financial statements.

According to Table 2, the evaluation of the data presented in this work based on selected features using the squirrel optimization algorithm and the closest neighbor classification algorithm has led to better performance and higher evaluation values than other classification algorithms.

**Table. 2** The averages of the evaluation criteria for the technique that was suggested, which is based on classification algorithms

|  | **Accuracy** | **Recall** | **Precision** |
|---|---|---|---|
| SSA_KNN | 98.8 | 98.54 | 98.97 |
| KNN | 96.64 | 94.71 | 93.06 |
| NN | 70.70 | 68.98 | 79.98 |
| NB | 96.41 | 93.94 | 93.82 |

## 5. Conclusion

One of the most widespread ways to undermine the integrity of the banking and financial systems is through the practice of fraudulent financial reporting. Monitoring all day-to-day transactions and following up on all instances of suspected fraud would involve a lot of staff, as well as a significant amount of time and money. This will also do significant harm to the financial and banking systems. Therefore, over the last ten years, academics have devised ways to provide approaches that are more cost-effective by analyzing the features of transactions and developing models for detecting financial statement fraud. Data mining methods play a crucial part in data mining and the recovery of tacit and secret data from vast amounts of data, which may lead to structures that are helpful for discovering and detecting fake financial statements. Due to the scarcity of transaction fraud, the issue of data instability might manifest itself in the form of finding anomalous samples in the relevant data set and recognizing financial statement fraud. In addition, the impossibility of tracing all financial statements to show their validity or fabrication has rendered a substantial number of the files undetermined, and only a tiny fraction of this data has been given, especially to regular and fraudulent consumers. In addition, the numerous and diverse components of financial statements have necessitated a reduction in size and the careful selection of just the most useful components. In this research, a method for identifying financial statement fraud based on a mix of attribute selection based on the squirrel optimization algorithm and classification algorithms is provided. Using feature selection based on the squirrel optimization algorithm and closest neighbor classification algorithm, the suggested technique has gotten better efficiency and increased evaluation rates than existing classification methods.